\documentclass[pmlr]{jmlr} 

\RequirePackage{graphicx}
 \usepackage{booktabs}
\usepackage{longtable}% for long tables
\usepackage{amsmath, amssymb} % math symbols
\usepackage[capitalise]{cleveref} % easier refs
\usepackage{overpic} % labeling panels with A, B, ...
\usepackage{subcaption}
\usepackage{multirow} % multirow tables
\usepackage{url}

\usepackage{soul,color}

\makeatletter
\def\set@curr@file#1{\def\@curr@file{#1}} %temp workaround for 2019 latex release
\makeatother
\usepackage[load-configurations=version-1]{siunitx} % newer version

\theorembodyfont{\upshape}
\theoremheaderfont{\scshape}
\theorempostheader{:}
\theoremsep{\newline}

\jmlrproceedings{PMLR}{Proceedings of Machine Learning Research}
\jmlrvolume{340}
\jmlryear{2026}
\jmlrworkshop{Machine Learning for Healthcare}

\title[A foundation-model approach to pediatric headache classification from rs-fMRI]{A foundation-model approach to pediatric headache classification from resting-state fMRI}

\newcommand{\equal}[1]{{\hypersetup{linkcolor=black}\thanks{#1}}}

\author{%
\Name{Guilherme S. {Imai Aldeia}}\Email{guilherme.imaialdeia@childrens.harvard.edu}\\
\addr Computational Health Informatics Program, Boston Children's Hospital, Boston, MA, USA
\addr Department of Pediatrics, Harvard Medical School, Boston, MA, USA
\AND
\Name{Clara Moon}\Email{clara.moon@childrens.harvard.edu}\\
\addr Pediatric Pain Pathway Lab;
\addr Department of Anesthesia, Critical Care and Pain Medicine;
\addr Department of Anesthesia, Boston Children's Hospital, Boston, MA, USA
\AND
\Name{Julie Shulman}\Email{julie.shulman@childrens.harvard.edu}\\
\addr Department of Anesthesia, Critical Care and Pain Medicine;
\addr Department of Anesthesia;\\
\addr Pediatric Pain Rehabilitation Center, Boston Children's Hospital, Boston, MA, USA
\AND
\Name{Navil Sethna}\Email{navil.sethna@childrens.harvard.edu}\\
\addr Department of Anesthesia, Critical Care and Pain Medicine;
\addr Department of Anesthesia;
\addr Pediatric Pain Rehabilitation Center, Boston Children's Hospital, Boston, MA, USA
\AND
\Name{Allison Smith}\Email{allison.smith@childrens.harvard.edu}\\
\addr Department of Anesthesia, Critical Care and Pain Medicine;
\addr Department of Anesthesia;
\addr Pediatric Headache Program, Boston Children's Hospital, Boston, MA, USA
\AND
\Name{Alyssa Lebel}$^*$\Email{alyssa.lebel@childrens.harvard.edu}\\
\addr Department of Anesthesia, Critical Care and Pain Medicine;
\addr Department of Anesthesia, Boston Children's Hospital, Boston, MA, USA
\AND
\Name{William G. {La Cava}}$^*$\Email{william.lacava@childrens.harvard.edu}\\
\addr Computational Health Informatics Program, Boston Children's Hospital, Boston, MA, USA;
\addr Department of Pediatrics, Harvard Medical School, Boston, MA, USA
\AND
\Name{Scott Holmes}\equal{Equal contribution}\Email{scott.holmes@childrens.harvard.edu}\\
\addr Pediatric Pain Pathway Lab;
\addr Department of Anesthesia, Critical Care and Pain Medicine;
\addr Department of Anesthesia, Boston Children's Hospital, Boston, MA, USA
}

\begin{document}

\maketitle

\begin{abstract}
    Headache is the most common neurological disorder in children and substantially affects quality of life.
    We investigated whether resting-state functional MRI (rs-fMRI) can support pediatric headache classification using machine learning.
    We encoded rs-fMRI data using NeuroSTORM, a recent foundation model, and fine-tuned it to distinguish healthy controls from children with headache and subsequently classify headache subtypes.
    We then compared NeuroSTORM with a standard neuroscience approach that uses functional connectivity (FC) matrices derived from brain activity as predictors.
    Using $189$ rs-fMRI scans from $110$ individuals collected across two visits (prevalence of any headache: 74\%), NeuroSTORM achieved an area under the receiver operating characteristic curve (AUROC) of $0.82$ ($95\%$ CI, $0.82$--$0.82$) and an area under the precision-recall curve (AUPRC) of $0.93$ ($95\%$ CI, $0.93$--$0.94$) in discriminating headache from non-headache.
    In contrast, models trained on FC matrices showed limited performance (AUROC, $0.67$ [$95\%$ CI, $0.67$--$0.67$]; AUPRC, $0.85$ [$95\%$ CI, $0.85$--$0.85$]).
    In a multiclass setting, when tasked with classifying individuals as healthy controls, individuals with chronic migraine, or individuals with non-chronic headaches (e.g., post-viral headache, new daily persistent headache, post-traumatic headache), NeuroSTORM achieved a macro-AUROC of $0.69$ ($95\%$ CI, $0.68$--$0.69$).
    The results suggest this approach can distinguish chronic migraine, the most common headache syndrome, but has difficulty differentiating other headache subtypes from chronic migraine.
    Overall, under limited-data conditions, NeuroSTORM appears to capture latent rs-fMRI representations that transfer to headache-related tasks.
    The findings provide proof-of-concept for fMRI-based prediction of pediatric headache using a foundation model without relying on functional connectivity features and highlight the potential of this approach for subtype identification.
    Further development of such tools may help clinicians improve diagnosis based on brain activity and tailor treatment strategies to individual patients.
\end{abstract}

\section{Introduction}
\label{introduction}

Headache is the most prevalent neurologic disorder in children, affecting at least $58.4\%$ of the pediatric and adolescent population~\citep{prezioso2022pediatric}.
Chronic daily headache impacts $1$--$3\%$ of adolescents globally~\citep{seshia2012chronic,cuvellier2008cephalees,arruda2010frequent}, with U.S. prevalence reaching $6\%$~\citep{pawlowski2019national}.
Among subtypes, chronic migraine is the most common and recurrent headache syndrome~\citep{https://doi.org/10.1111/dmcn.14338}, with a much higher prevalence in girls than in boys during adolescence~\citep{Blume2012}.
These conditions can severely impact daily life, leading to school absence, cognitive difficulties, nausea, and photophobia~\citep{langdon2017pediatric}.

Machine learning (ML) has become increasingly common in neuroimaging applications~\citep{pereira2009machine, khosla2019machine} as a data-driven approach for identifying complex patterns and supporting automated decision-making.
Brain connectivity plays a key role in such applications, commonly measured using functional Magnetic Resonance Imaging (fMRI), a non-invasive technique that captures hemodynamic responses associated with neural activity~\citep{logothetis2008we}, that is, the blood-oxygen-level dependent (BOLD) signal.
Both fMRI and its derived functional connectivity (FC) matrices---pairwise correlation coefficients between brain regions---can be used to investigate how brain function relates to disorders, with FC often preferred because of its lower dimensionality and more interpretable features.

When acquired during resting state (rs-fMRI), these patterns reflect the brain's intrinsic organization~\citep{varoquaux2013learning} and provide a valuable tool for investigating neurological disorders.
Prior work in cases such as post-traumatic headache has shown the potential for structural MRI and FC matrices to be used in the context of differentiating patient cohorts~\citep{dumkrieger_value_2023,joshi_prediction_2024,chong_migraine_2017,mao_identifying_2025}.
Relatively few studies have explored FC-based analyses in pediatric cohorts~\citep{holmes_integrated_2022,Ofoghi2021}.
Such data-oriented approaches are particularly relevant in pediatric populations because young children often struggle to describe their symptoms~\citep{LANGDON201744}, and diagnosis relies on semi-structured interviews and following the ICHD-3 \citep{ichd3_2018} guidelines. Disability and quality of life can be assessed using questionnaires such as the HIT-6~\citep{RendasBaum2014}. When neuroimaging is used for diagnosis, it primarily focuses on investigating potential structural characteristics, such as abnormalities, inflammation, and ischemia~\citep{Blume2012}.

Despite advances in ML, fMRI-based applications still lack standardized pipelines and strong transferability.
A key reason is that many brain-imaging studies are not replicable~\citep{poldrack2017transparent}, involve too few participants to support robust conclusions~\citep{button2013power,marek2022reproducible}, and brain structures exhibit substantial inter-individual differences~\citep{brett_problem_2002}.
Recent work has explored architectures such as the Swin Transformer~\citep{liu_swin_2021} and the Mamba state space model~\citep{gu_mamba_2024} for fMRI data, and the emergence of foundation models trained on more than $28$ million fMRI frames~\citep{wang_towards_2025} has shown that direct processing of fMRI images can generate effective representations.
However, these studies have been limited to adult populations and applications to pediatric health remain limited.
Furthermore, the application of foundation models to fMRI remains a very recent development in the field.

Here, we investigate whether ML models can classify pediatric headache using either FC features or direct fMRI data encoded with a foundation model. 
We collected data from $110$ individuals across two waves, totaling $189$ scans from participants aged $8$--$22$ years at the time of visit. 
Participants were either healthy controls ($n=45$ individuals) or had a headache condition, with chronic migraine being the most frequent subtype ($n=68$ individuals). 
We preprocessed the fMRI data, extracted FC matrices, and trained shallow ML methods to classify pediatric headache. 
We then trained a final model to differentiate headache subtypes based on the best-performing binary-classification setting.

\subsection*{Generalizable Insights about Machine Learning in the Context of Healthcare}

% This section is \emph{required}, must keep the above title, and should be the final part of your introduction.  In about one paragraph, or 2-4 bullet points, explain what we should \emph{learn} from reading this paper that might be relevant to other machine learning in health endeavors.

% For example, a work that simply applies a bunch of existing algorithms to a new domain may be useful clinically but doesn't increase our understanding of the machine learning and healthcare; if that study also investigates \emph{why} different approaches have different performance, that might get us excited!  A more theoretical machine learning work may be in how it enables a new kind of clinical study. \emph{Reviewers and readers will look to evaluate (a) the significance of your claimed insights and (b) evidence you provide later in the work of you achieving that contribution}

This work presents the first proof-of-concept of effective pediatric headache classification from rs-fMRI data. 
The results are made possible by the novel application of a pre-trained foundation model (NeuroSTORM) to directly encode fMRI data, which outperformed traditional FC-based features in a limited-data setting.
More broadly, it suggests that the recent emergence of large-scale, pre-trained neuroimaging models provides a practical starting point for clinical ML applications in which site-specific datasets are too small to support training high-capacity models from scratch.
At the same time, our results highlight the need for careful evaluation of subgroup performance, interpretability, and prospective validation before such methods can be used as clinical decision-support tools.

\section{Related Work}

% Make sure you also put your work in the context of related work. Who else has worked on this problem, and how did they approach it?  What makes your direction interesting or distinct?

Prior work has leveraged rs-fMRI and FC for migraine classification in adults ($58$ migraine individuals and $50$ healthy controls), applying ML to distinguish individuals using $33$ brain regions manually pre-selected based on findings from the pain and migraine literature~\citep{chong_migraine_2017}. The authors reported that migraine is easier to classify in individuals older than $14$ years, with classification accuracy decreasing from $96\%$ to $82\%$ for individuals aged $14$ years or younger.

In a related line of work,~\cite{holmes_integrated_2022} investigated post-traumatic headache (PTH) in a pediatric and young cohort ($64$ PTH individuals and $34$ healthy controls), a condition with still unclear underlying mechanisms. Using ML techniques combined with data reduction, feature selection, and clustering, the authors reduced $70$ regions of interest to $14$ salient features. Similarly,~\cite{mao_identifying_2025} studied PTH in adults ($73$ individuals), aiming to predict clinical improvement three months after mild traumatic brain injury (mTBI). Complementary,~\cite{Ofoghi2021} studied PTH after mTBI in children, using FC-based features and general linear models. ~\cite{qiu_mapping_2023} examined new daily persistent headache (NDPH) ($35$ individuals with NDPH and $40$ healthy controls) from a functional connectivity perspective using magnetoencephalography in an adult cohort.
These studies predominantly focus on adult populations and typically rely on derived FC representations rather than direct rs-fMRI images.

More recently, foundation models have begun to be applied across healthcare domains. These models are trained on large-scale, heterogeneous datasets and learn generalizable representations of the data of interest --- in this case, fMRI.
In neuroscience, this paradigm has led to models such as fMRI-LM, which embeds fMRI-derived tokens into a language space to enable high-level, natural language interpretation of low-level neuroimaging signals~\citep{wei_fmri-lm_2025}. Another example is SwiFT~\citep{kim_swift_2023}, a model based on the Swin Transformer~\citep{liu_swin_2021} that learns brain dynamics directly from fMRI volumes. SwiFT was trained on large cohorts, including some of the largest publicly available datasets: the Human Connectome Project (HCP)~\citep{VanEssen2012}, Adolescent Brain Cognitive Development (ABCD)~\citep{Casey2018}, and UK Biobank (UKB)~\citep{Sudlow2015}; and validated on downstream tasks such as brain age prediction. More recently, NeuroSTORM~\citep{wang_towards_2025} was proposed and trained over more than $28$ million fMRI frames, replacing the standard attention mechanism with a state space model (Mamba)~\citep{gu_mamba_2024}, and demonstrating strong performance across phenotype and diagnosis prediction tasks.

In this work, we bridge the gap between prior ML-based approaches for headache prediction and the emerging use of foundation models for fMRI headache prediction in pediatric and youth populations. Leveraging a sample size of $189$ scans, we utilize the encoder of the pre-trained NeuroSTORM model to predict pediatric headache outcomes, and contrast its performance with traditional ML approaches commonly used in the neuroscience literature.

\section{Materials and methods}

We tackle two classification tasks: (1) binary classification to distinguish healthy controls from pediatric headache patients, and (2) multi-class classification to distinguish healthy controls, chronic migraine, and non-chronic migraine using a one-vs-all scheme.
We focus on differentiating chronic migraine because it is the most common and well-defined headache subtype. Other headache subtypes are more susceptible to labeling noise because their diagnostic criteria are less well defined and their symptoms often overlap.

\begin{figure}[tbh]
    \centering
    \includegraphics[width=\linewidth,clip=true,trim={0.8cm 0.5cm 0.7cm 0.3cm}]{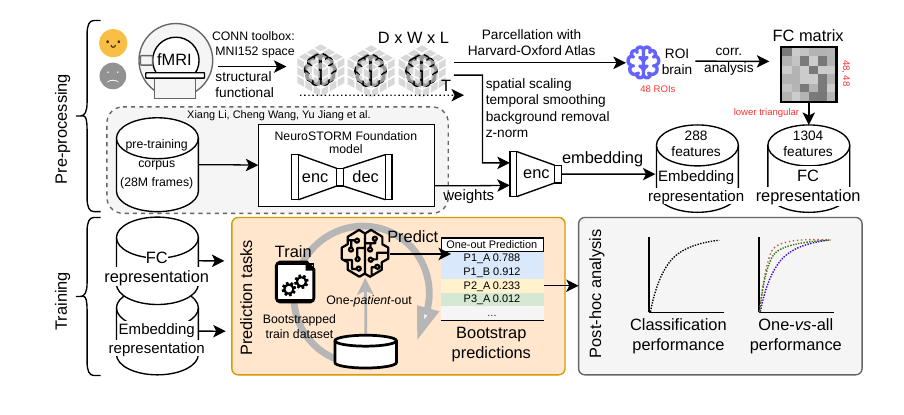}
    \caption{
        Experimental pipeline.
        The data are preprocessed to generate two distinct representations, one using features extracted from functional connectivity matrices and the other using embeddings extracted directly from the fMRI data using a foundation model.
        Both feature representations are used as inputs for training machine learning methods using a bootstrapped leave-one-out approach.
    }
    \label{fig:experiments-pipeline}
\end{figure}

We use two representations to compare approaches: one uses traditional FC matrices derived from cortical brain regions, while the other uses embeddings from a foundation model (NeuroSTORM) applied directly to fMRI data. This comparison allows us to evaluate whether the foundation model outperforms the standard approach and to quantify how well traditional FC-based methods perform on this headache-classification problem.
Figure~\ref{fig:experiments-pipeline} illustrates the overall experimental pipeline.

The raw fMRI data were first preprocessed using the CONN toolbox~\citep{WhitfieldGabrieli2012} functional preprocessing pipeline with default parameters and then mapped to MNI152 space, followed by the default denoising pipeline.
Mapping to MNI152 space is common in neuroscience and was also used by the NeuroSTORM authors when preparing the training data for the foundation model.
The CONN default pipeline includes brain extraction, functional realignment and normalization, slice-timing correction, outlier identification, smoothing with a kernel of $8\,\mathrm{mm}$ FWHM, and bandpass filtering between $[0.008\,\mathrm{Hz}, 0.09\,\mathrm{Hz}]$ to remove BOLD fluctuations that may be generated by physiological processes, head motion, and other noise sources.
Brain extraction helps remove confounding signal outside the brain, and smoothing increases the signal-to-noise ratio in the time series.
The resulting data were used to generate FC and embedding representations

For the FC representation, the Harvard--Oxford atlas is used for parcellation, containing $48$ cortical and $21$ subcortical regions of interest (ROIs).
Given the sample size, we kept only the cortical ROIs to obtain a more favorable feature-to-sample ratio.
We chose not to perform feature selection on the FC features. Although feature selection may increase statistical power, it can produce cohort-dependent feature sets and reduce generalizability. Pairwise correlations were computed between all ROI time series, yielding a symmetric $48 \times 48$ functional connectivity matrix.
The upper triangular portion and main diagonal were discarded, resulting in $1,128$ unique connectivity features. Including subcortical regions would increase the feature space to $2,346$ features.

For the embedding representation, the rs-fMRI volumes undergo the preprocessing proposed by the NeuroSTORM authors, with spatial scaling to fit the default input size of $96 \times 96 \times 96$ and temporal smoothing to match the input requirement of the pre-trained foundation model ($20$ time frames).
Background voxels are removed, and z-normalization is applied before passing the data through the encoder, yielding a $288$-dimensional feature embedding.

\begin{figure}[tbh]
    \centering
    \includegraphics[width=\linewidth,clip=true,trim={1cm 0 0 1cm}]{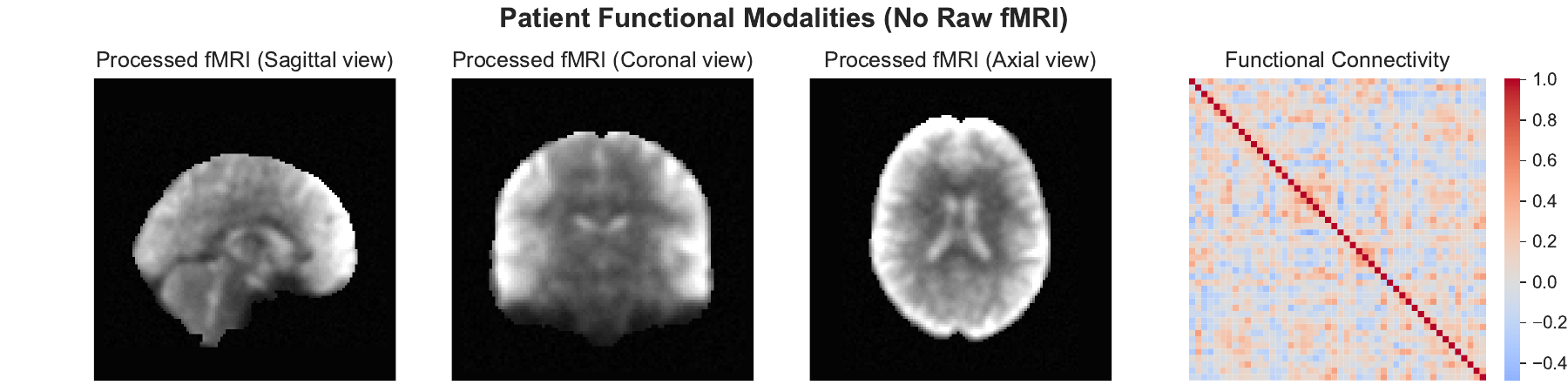}
    \caption{
        Illustrative example of the data used in the experiments.
        The first three images show three views of a 4-dimensional image after applying the preprocessing pipeline, which includes denoising, MNI152 space alignment, and smoothing.
        The last image shows a functional connectivity matrix depicting pairwise correlations among $48$ ROIs over time.
    }
    \label{fig:imaging_examples}
\end{figure}

Figure~\ref{fig:imaging_examples} illustrates the data modalities used in this study. We do not use raw fMRI directly because of its low signal-to-noise ratio and because it contains structural information unrelated to the brain (e.g., eyes and other tissues). Regardless of whether one uses fMRI or FC representation, standard preprocessing is routine in the neuroscience literature before model development.

Using these two representations, we train a range of shallow machine learning classifiers to distinguish between healthy controls and headache participants. Given the limited sample size, we use leave-one-out (LOO) cross-validation, in which each individual is held out once while the model is trained on the remaining participants. When an individual is held out for testing, all of that participant's visits are excluded from the training data to avoid leakage. 
The intent of this approach is to use all available data while ensuring that no visit from a test participant appears in the training set.
This LOO procedure is repeated $100$ times with different random seeds to capture uncertainty in the model's predicted probabilities for each participant.

After identifying the best-performing model for the binary classification task, a final experiment is conducted to predict headache subtypes. This stage follows the same LOO cross-validation and bootstrapping procedure used for the first experiment, in a one-vs-all schema.

\subsection{Eligibility Criteria and Recruitment}

A total of $110$ participants were recruited from the Greater Boston area. Data were collected in two waves, yielding $189$ scans: $104$ acquired at the first visit and $85$ acquired at the second visit. $79$ participants completed both visits.
Of the $189$ scans, $144$ were acquired from patients with a formal headache diagnosis who were evaluated through Boston Children’s Hospital’s Chronic Headache Program, Sports Medicine Clinic, Adolescent/Young Adult Medicine, Pediatric or Young Adult Pain Rehabilitation Centers, or Post-Acute Sequelae of SARS-CoV-2 clinic.
Following diagnosis, eligible patients were invited to participate during or after their clinical visits.

Eligibility was determined through pre-screening questionnaires that assessed headache history and screened for exclusion criteria, including concurrent pain conditions and mental health disorders. Participants were eligible if they were $8$--$22$ years old; of any gender, including non-binary and transgender; able to speak and read English sufficiently; and free of other acute or chronic medical or pain conditions that could confound the data. Additional exclusion criteria included metallic implants, pregnancy, and claustrophobia. Healthy controls ($n=45$) were pain-free individuals from the hospital or the local community with no history of headache, recruited through advertisements, fliers, the hospital's internal website, and word of mouth.

All persons with a headache were evaluated clinically with oversight by study physician (AL). All persons were evaluated within the Boston Children’s Hospital system with clinical mention of headache as their primary (e.g., chronic migraine, new daily persistent headache), or secondary (e.g., post-traumatic headache, post-viral headache) symptoms. 

The Boston Children's Hospital Institutional Review Board approved the data-acquisition protocol used in this study and waived the requirement for informed consent.

\subsection{Imaging}

Brain imaging was acquired on a $3$T Siemens Prisma scanner equipped with a $32$-channel phased-array head coil.
High-resolution T1-weighted magnetization-prepared rapid acquisition gradient echo (MPRAGE) images were acquired. 
Resting-state images were acquired using the following parameters: isotropic voxel size $= 2.4 \times 2.4 \times 2.4\,\mathrm{mm}^3$, TR $= 3420\,\mathrm{ms}$, TE $= 33.0\,\mathrm{ms}$, and slice thickness $= 2.4\,\mathrm{mm}$.

\subsection{Clinical and Demographic Information}~\label{subsec:demographic}

Participants completed demographic questionnaires reporting age, assigned sex at birth, and years of education, along with a brief clinical assessment battery.
This battery included the Head Impact Test-6 (HIT-6), which evaluates headache symptoms and severity, and the Central Sensitization Inventory (CSI), which assesses symptoms related to central sensitization and pain amplification in the central nervous system.
The HIT-6 score can range from $36$ to $78$, with higher scores indicating greater impact on daily life due to headache.
The CSI can range from $0$ to $100$ and uses a Likert scale with responses from $0$ (never) to $4$ (always) to score symptoms commonly associated with central sensitization syndrome.
Demographic information is summarized in Table~\ref{tab:statistics-data}. 

\begin{table}[tbh]
    \caption{
        Demographic statistics of the study population.
        Values are reported as mean (SD) unless otherwise specified.
        HIT-6 and CSI scores are reported by cohort only.
    }
    \label{tab:statistics-data}
    \centering
    \small
    \begin{tabular}{@{}rrrrr@{}}
    \toprule \midrule
                      &                &             & \multicolumn{2}{c}{Cohort}    \\ \cline{4-5} 
                      &                & Overall     & Healthy      & Headache       \\ \midrule
    \multicolumn{2}{r}{n, counts (\%)}       & 189       & 45 (23.8)    & 144 (76.2)  \\ \midrule
    Assigned sex                 & F, counts (\%)  & 133 (70.4)   & 21 (46.7)    & 112 (77.8)   \\
    at birth                    & M, counts (\%)  & 56 (29.6)   & 24 (53.3)    & 32 (22.2)   \\ \midrule
    \multicolumn{2}{r}{Age} & 15.7 (2.8)  & 15.5 (4.2)   & 15.7 (2.3)  \\ \midrule
    \multicolumn{2}{r}{HIT-6} & -- & 44.0 (6.1) & 61.3 (7.0)  \\ \midrule
    \multicolumn{2}{r}{CSI} & -- & 18.6 (15.6)  & 43.9 (16.7) \\ \midrule \bottomrule
    \end{tabular}
\end{table}

\begin{figure}[tbh]
    \centering
    \begin{minipage}[t]{0.33\textwidth}
        \centering
        \includegraphics[width=\linewidth]{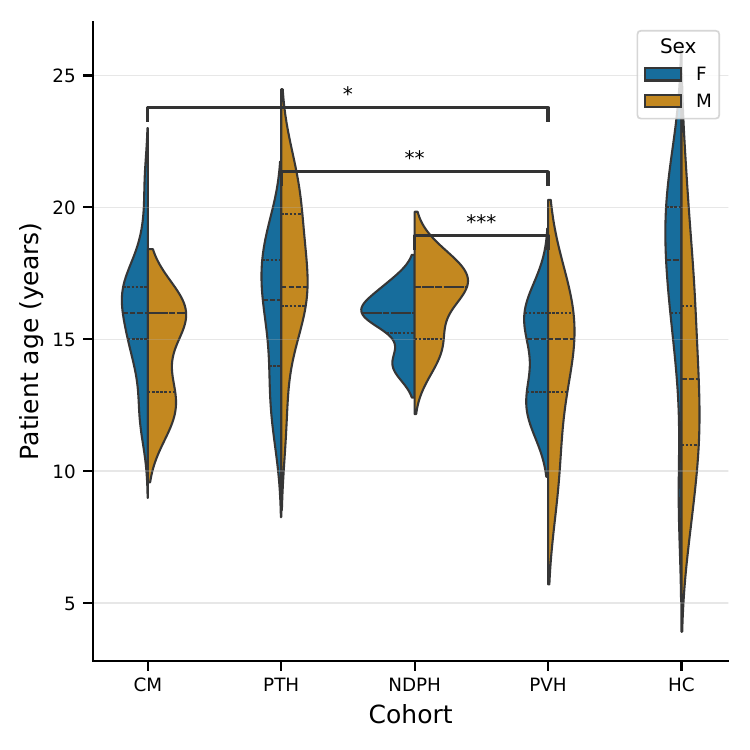}
        \\[5pt]
        {\small \textbf{A} Age distribution}
        \label{fig:age_stats}
    \end{minipage}%
    \hfill
    \begin{minipage}[t]{0.33\textwidth}
        \centering
        \includegraphics[width=\linewidth]{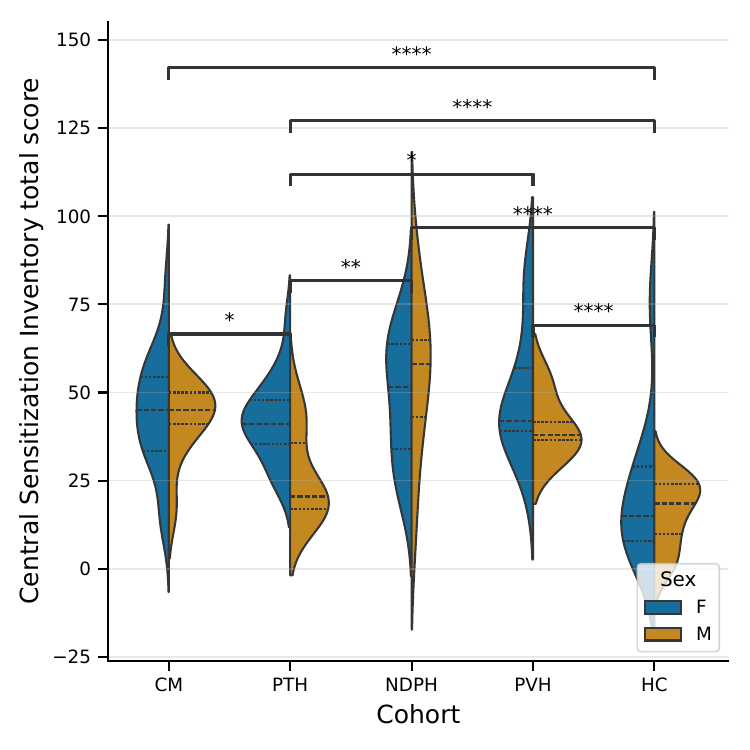}
        \\[5pt]
        {\small \textbf{B} CSI scores}
        \label{fig:csi_stats}
    \end{minipage}%
    \hfill
    \begin{minipage}[t]{0.33\textwidth}
        \centering
        \includegraphics[width=\linewidth]{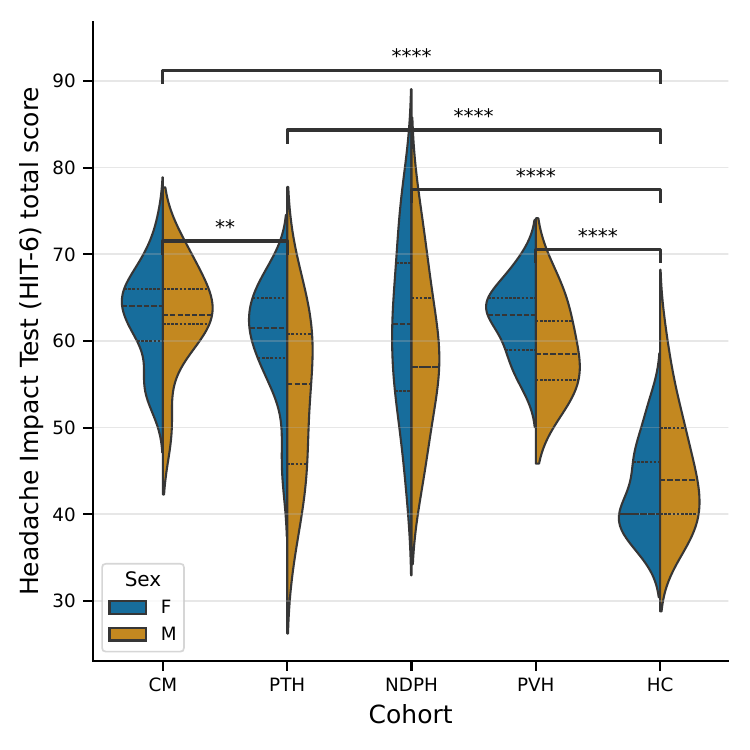}
        \\[5pt]
        {\small \textbf{C} HIT-6 scores}
        \label{fig:hit6_stats}
    \end{minipage}
    \caption{
        Clinical and demographic measurements by cohort.
        (A) Age distribution across healthy controls and headache patients.
        (B) Central Sensitization Inventory (CSI) scores.
        (C) Head Impact Test-6 (HIT-6) scores.
        $^{*}$ $10^{-2} < p \leq 5\times10^{-2}$,
        $^{**}$ $10^{-3} < p \leq 10^{-2}$,
        $^{***}$ $10^{-4} < p \leq 10^{-3}$,
        $^{****}$ $p \leq 10^{-4}$.
        non-significant differences are omitted.
    }
    \label{fig:violin-plot-stats}
\end{figure}

Distributions of age, CSI scores, and HIT-6 scores for healthy controls (HC, n=$45$, $23.81\%$) and the headache subtypes are presented in Figure~\ref{fig:violin-plot-stats}, split by assigned sex at birth.
In this figure, statistical tests were performed between the HC group and each headache subtype, as well as pairwise comparisons between each headache subtype. Bonferroni correction was applied. Non-significant results were omitted.

The final cohort included data from patients with chronic migraine (CM; $35.98\%$, $68$ of $189$ scans) and other identified headache subtypes ($40.21\%$, $76$ of $189$ scans), including post-viral headache (PVH), new daily persistent headache (NDPH), and post-traumatic headache (PTH).

We first trained a demographic logistic regression model using only age and assigned sex at birth to assess whether performance in the binary task could be explained by cohort composition rather than neuroimaging signal.
This comparison is important because the cohorts differ in sex distribution, while age differences are modest, making sex a potential confounder.
Age was discretized into six bins: $18$ months--$3$ years, $3$--$5$ years, $5$--$10$ years, $10$--$15$ years, $15$--$18$ years, and above $18$ years. We then fit an unregularized logistic regression model using discretized age and sex as demographic predictors.

\subsection{Machine learning algorithms}~\label{subsec:ml}

We evaluated a range of ML methods, considering both interpretable models and black-box approaches. Interpretable models included decision trees (DT) and logistic regression (LR), while black-box models comprised ensemble methods, namely random forests (RF) and gradient boosting (GB). All models were implemented in scikit-learn v1.8.0.

Preliminary experiments were conducted to constrain model complexity for the FC-based approach, whose features are clinically meaningful, while avoiding extensive tuning of the foundation-model pipeline. Given the limited sample size and the risk of overfitting, we intentionally avoided extensive hyperparameter optimization. Instead, we restricted tuning primarily to parameters that directly affect model complexity, and primarily relied on standard configurations.

Performance degraded for DT when attempting to limit the maximum depth to $2$, $3$, or $5$, and thus we retained the default hyperparameters. Similarly, reducing the number of estimators in RF and GB below the default value of $100$ also resulted in worse performance.
We tested $100$, $1000$, and $5000$ maximum iterations for logistic regression methods. Increasing maximum iterations from $100$ to $1000$ improved performance, but not beyond $1000$. A detailed description of the model configurations used to obtain the reported results is provided in Appendix \ref{apx:methods}.

Finally, because embedding encoders are designed to produce linearly separable representations, we focus on logistic regression with L1 regularization for the embedding-based representation rather than more complex models such as neural networks.

\section{Results}

Group differences were assessed using Welch's two-sample $t$-test.
Consistent with the descriptive statistics in Table~\ref{tab:statistics-data}, participants with headache exhibited substantially higher HIT-6 scores (headache: $61.3 \pm 7.0$; healthy: $44.0 \pm 6.1$), a difference that was highly significant ($t = 14.84$, $p = 2.00 \times 10^{-33}$).
CSI scores were also significantly elevated in the headache group (headache: $43.9 \pm 16.7$; healthy: $18.6 \pm 15.6$; $t = 9.02$, $p = 2.19 \times 10^{-16}$).
Mean age differed modestly between cohorts (headache: $15.7 \pm 2.3$; healthy: $15.5 \pm 4.2$), but this difference was not statistically significant ($t = 0.43$, $p = 0.68$).

Classification performance is summarized in Table~\ref{tab:supervised_metrics}.
For each metric, we report the mean performance along with bootstrapped $95\%$ confidence intervals.

\begin{table}[tbh]
    \caption{
        Classification performance for the models evaluated with functional connectivity (DT, LR L1/L2, GB, and RF) and embedding-based features (NeuroSTORM).
        AUROC: Area Under Receiver Operating Characteristic Curve;
        AUPRC: Area Under Precision-Recall Curve.
        Values are reported as (mean, 95\% CI).
        The highlighted value in each column depicts the best performing model for that specific measure.
    }
    \label{tab:supervised_metrics}
    \centering
    \small
    \begin{tabular}{@{}rrr@{}}
    \toprule\midrule
    & AUROC & AUPRC \\ \midrule
    Demographic Model & $0.61\ [0.61, 0.61]$ & $0.82\ [0.80, 0.83]$ \\
    DT & $0.53\ [0.48, 0.56]$ & $0.77\ [0.75, 0.79]$ \\
    LR L1 & $0.67\ [0.67, 0.67]$ & $0.85\ [0.85, 0.85]$ \\
    LR L2 & $0.58\ [0.58, 0.58]$ & $0.79\ [0.79, 0.79]$ \\
    GB & $0.40\ [0.38, 0.43]$ & $0.71\ [0.69, 0.73]$ \\
    RF & $0.54\ [0.50, 0.57]$ & $0.78\ [0.75, 0.80]$ \\
    NeuroSTORM & $\mathbf{0.82\ [0.82, 0.82]}$ & $\mathbf{0.93\ [0.93, 0.94]}$ \\ \midrule\bottomrule
    \end{tabular}
\end{table}

Among the FC-based classifiers, decision trees yielded low performance (AUROC $=0.53$, AUPRC $=0.77$), close to a random classifier.
Ensemble methods (i.e., GB, RF) did not outperform linear models, with the best-performing ensemble achieving AUROC $= 0.54$ and AUPRC $= 0.78$.
Logistic regression with L1 and L2 regularization exhibited very tight confidence intervals, with variations detected only at the fourth decimal place. 

The best performance was observed for NeuroSTORM, which uses logistic regression with L1 regularization on the embedding representation, achieving an AUROC of $0.82$ and an AUPRC of $0.93$. These results were robust to variation in the random seed, with no observable differences within the $95\%$ confidence interval. The strong performance of a simple linear classifier on the embeddings suggests that the foundation model captures resting-state dynamics in a representation that transfers well to this downstream task.
More broadly, the narrow confidence intervals suggest that, in this experimental setting, LR converges to a highly similar solution across seeds despite the stochastic solver.

Given the poor performance from the GB model, we inspected the training logs in detail and noticed that AUROC training performance was always equal to one. Inspecting feature importance obtained with SHAP \citep{NIPS2017_8a20a862} explanations across runs, we additionally noted that there was large variation in the most important features. In this light, we believe the model is overfitting to spurious correlations, resulting in an LOO performance worse than guessing.

The demographic logistic regression model for predicting any headache achieved an AUROC of $0.61$ and an AUPRC of $0.82$. Coefficient inspection showed that non-zero weights were concentrated in age-bin indicators.
To evaluate the robustness of different models to potential demographic confounding, we performed a stratified analysis by sex and computed AUROC and reported the results in Table \ref{tab:supervised_metrics_stratified}. Notably, NeuroSTORM maintained performance across demographic groups, whereas the FC-based approaches exhibited a greater degree of performance degradation. These results suggest that performance is not due to confounding by sex.

\begin{table}[tbh]
    \caption{
        AUROC performance for the models when stratified by assigned sex at birth.
        Values are reported as (mean, 95\% CI).
        The highlighted value in each column depicts the best performing model for that specific measure.
    }
    \label{tab:supervised_metrics_stratified}
    \centering
    \small
    \begin{tabular}{@{}rrr@{}}
    \toprule\midrule
          & Male                           & Female                         \\ \midrule
    DT         & $0.55\ [0.49, 0.62]$           & $0.50\ [0.45, 0.55]$               \\
    LR L1      & $0.71\ [0.71, 0.71]$           & $0.60\ [0.60, 0.60]$               \\
    LR L2      & $0.59\ [0.59, 0.59]$           & $0.51\ [0.51, 0.51]$               \\
    GB         & $0.38\ [0.35, 0.44]$           & $0.36\ [0.33, 0.38]$               \\
    RF         & $0.58\ [0.51, 0.65]$           & $0.45\ [0.39, 0.51]$               \\
    NeuroSTORM & $\mathbf{0.80\ [0.79, 0.81]}$           & $\mathbf{0.81\ [0.80, 0.81]}$               \\
    \midrule\bottomrule
    \end{tabular}
\end{table}

\begin{figure}[tbh]
    \centering
    \includegraphics[width=\linewidth]{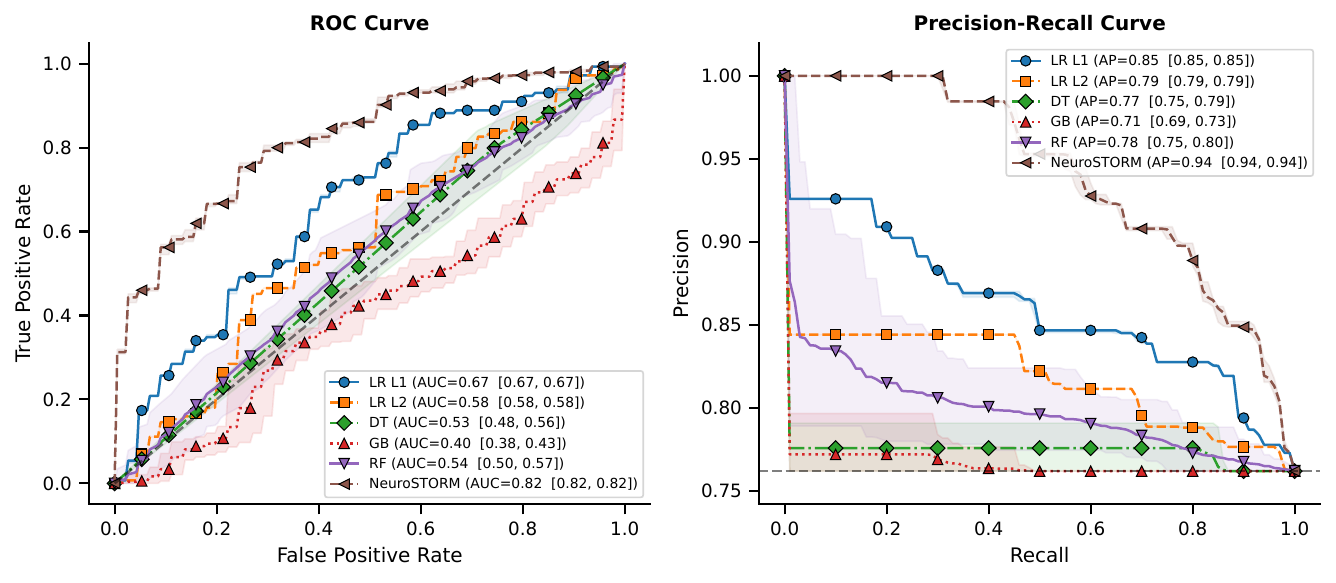}
    \caption{
        Precision-Recall and Receiver Operating Characteristic (ROC) Curves for the ML methods evaluated in the experiments, with the Area Under the Curve (AUC) and Average Precision (AP) reported for each curve. Confidence intervals are estimated using the leave-one-out evaluation procedure repeated $100$ times with different random seeds.
    }
    \label{fig:auprc_auroc}
\end{figure}

Precision--recall and ROC curves are presented in Figure~\ref{fig:auprc_auroc}. All models trained on the FC representation exhibit performance close to that of a random classifier (the diagonal dashed gray line in the ROC plot). We note that, because these curves are estimated using leave-one-out (LOO) validation on a limited dataset, individual points may correspond to predictions generated by different model instances rather than a single, fixed model.

\begin{figure}[tbh]
    \centering
    \includegraphics[width=\linewidth]{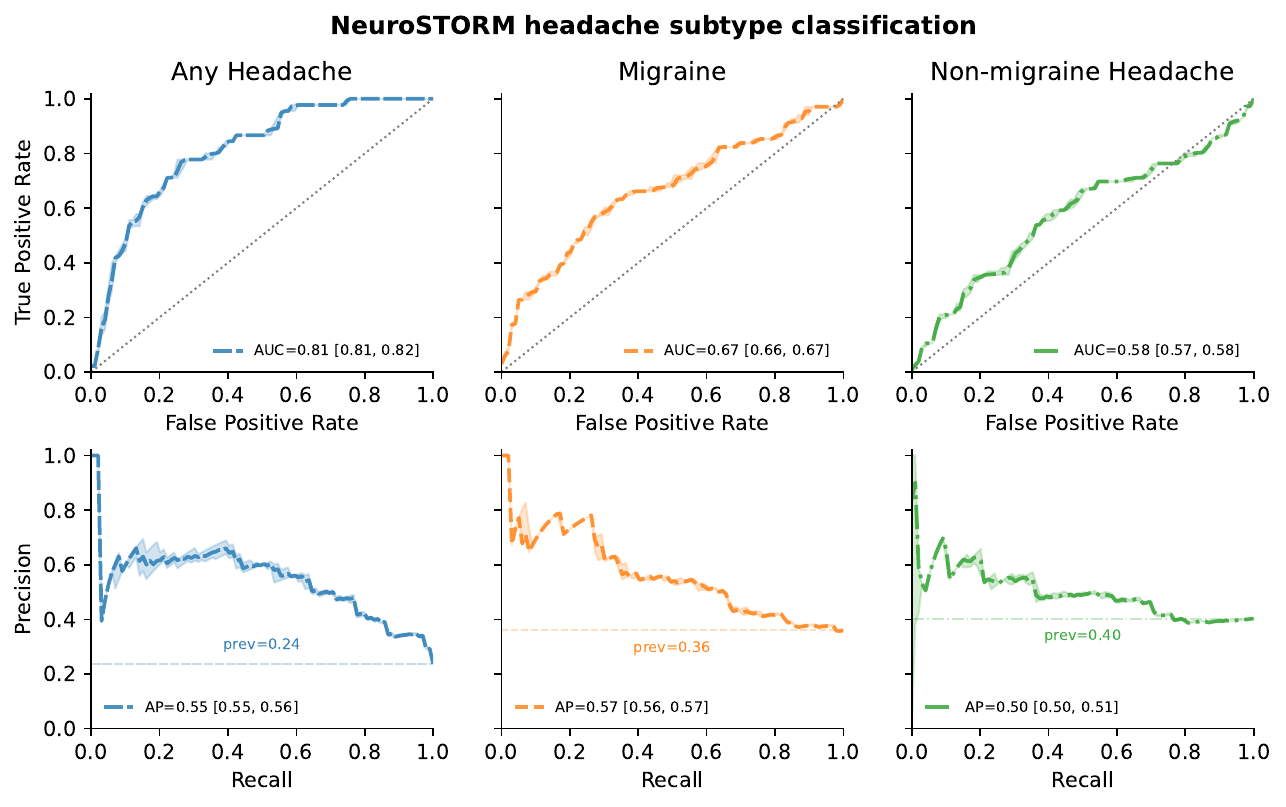} %,trim={0 0 0 1cm},clip=true
    \caption{
        Receiver Operating Characteristic (ROC) and Precision-Recall (PR) Curves for the NeuroSTORM method for predicting headache subtypes. The top row shows the ROC curves, with the Area Under the Curve (AUC) reported for each curve, while the bottom row shows the PR curves, with Average Precision (AP) reported for each curve
    }
    \label{fig:subtype_classification}
\end{figure}

Finally, we trained the best-performing model from the binary classification task to predict headache subtypes using the same experimental pipeline. Receiver Operating Characteristic and Precision-recall curves are presented in Figure~\ref{fig:subtype_classification}. The model achieved a macro-AUROC of $0.69$ ($95\%$ CI, $0.68-0.69$) and a weighted AUROC $= 0.66$ ($95\%$ CI, $0.66-0.67$). Overall, performance was lower for subtype discrimination than for binary classification, indicating that this is a more challenging task.

The bottom panels show the detailed precision--recall curves.
We observe that Migraine classification drive most of the overall performance, whereas the non-migraine types are less separable.
The classifier performed best for healthy controls (AUROC $=0.81$; AUPRC $= 0.55$, prevalence $0.24$) and chronic migraine (AUROC $= 0.67$; AUPRC $= 0.57$, prevalence $0.36$), but with a decreased performance for other headache (AUROC $= 0.58$; AUPRC $= 0.50$, prevalence $0.40$).

\section{Discussion} 

% \emph{This is probably the most important section of your paper! This is where you tell us how your work advances our understanding of machine learning and healthcare.}  Discuss both technical and clinical implications, as appropriate\cite{xyz19}.

% Discuss optimistic performance in binary task, and try to reason why it failed on subtypes and how it could be improved (e.g. the foundation model was not trained with headache data; data is limited and has different distributions; see literature)

Machine learning algorithms can help identify imaging patterns that differentiate pediatric patients with headache from those without a clinical headache history.
To date, however, most ML studies in neuroimaging have focused on cohort discrimination in adult populations and have primarily relied on seed-based FC matrices as input features.
In the current study, we evaluate several ML algorithms for differentiating pediatric patients with headache from healthy controls and compare FC features with foundation model embeddings

Headache and its subtypes remain challenging to diagnose, as they rely heavily on subjective self-report and criteria that are often difficult to operationalize, particularly in pediatric populations. 
Using HIT-6 and CSI scores, we confirmed that participants in the headache cohort exhibited elevated headache burden and symptoms consistent with central sensitization, suggesting that our participants experienced headache symptoms and had a significant centralized component.
Together with previous findings from~\cite{holmes_integrated_2022,holmes_disentangling_2023}, these studies highlight the value of integrating neuroimaging with clinical assessments to obtain objective markers of headache-related pathology.

In this work, we show that a foundation model used to encode pediatric rs-fMRI data can identify headache status with stronger discriminative performance than traditional methods based on FC matrices. We further compare this approach with standard FC-based methods without incorporating prior domain knowledge to restrict the feature space used for training. Although FC-based methods underperformed in our experiments, they offer clinically interpretable features, in contrast to the NeuroSTORM embedding representation, which lacks direct interpretability.

Ultimately, the value of ML and AI models in clinical settings depends on their ability to produce outputs that are clinically interpretable and actionable.
While the embeddings are not directly interpretable, future work can adapt tools such as GradCAM~\citep{jacobgilpytorchcam, Selvaraju_2019} for generating heatmaps to identify regions the foundation model focuses on for generating the encoded representation of the input fMRI.

The limited subtype performance may partly reflect the training data used for the foundation model, which predominantly consist of large cohorts of healthy individuals and, among disorder cohorts, populations such as pediatric ADHD (e.g., ADHD200), early psychosis, and other psychiatric diagnoses. As a result, the learned representations may transfer imperfectly to headache-specific outcomes. In addition, survey-based measures such as CSI and HIT-6 are inherently noisy because they depend on individual symptom perception and reporting.

When examining headache subtypes,~\cite{chong_migraine_2017} found that chronic migraine individuals with longer disease durations were more accurately classified. They also observed a marked effect of age: classification accuracy was substantially higher for individuals older than $14$ years compared to younger patients, with performance dropping by $14\%$ in the younger age group. Notably, their cohort had a mean age of $36.3$ years, whereas our pediatric cohort averages $15.7$ years.

The demographic model achieved an AUROC of $0.61$ and an AUPRC of $0.82$, indicating that age provide non-trivial discriminative signal even in the absence of neuroimaging features.
The imaging-based representation provides substantial additional predictive value in this cohort, as evidenced by the NeuroSTORM-based model, which achieved an AUROC of $0.82$ and an AUPRC of $0.93$.

The challenging performance for subtype classification, such as post-traumatic hedaches (PTH), may reflect inherent diagnostic and phenotypic complexities. Prior work~\citep{dumkrieger_value_2023} has argued that migraine and PTH exhibit similar neurobiological phenotypes, which could explain the lower discriminability between these subtypes. Notably, that study required extensive FC-based feature pre-selection (reducing $5,624$ features to $1,521$), suggesting that feature engineering may be a critical factor for distinguishing these phenotypically similar conditions. In our work, the superior performance for chronic migraine and lower performance for non-migraine headache may also be influenced by the nature of labeling: chronic migraine, being more well-defined clinically, may be more consistently labeled, while other diagnoses may carry greater uncertainty. This measurement variability in the target labels could contribute to differential classification performance across headache subtypes, given the NeuroSTORM performance drop from binary classification to subtype classification.

This study demonstrates the feasibility of leveraging modern AI methods to handle high-dimensional neuroimaging data such as fMRI.
Although fMRI is resource intensive, it provides additional information about the temporal dynamics of the brain that cannot be captured by measures used in standard of care.
As an initial step, the results support continued efforts to expand data collection, increase sample size, and adapt pre-trained models more directly to headache-related tasks.

Another contribution of this work is the extension beyond binary classification to headache subtype identification, a more challenging and clinically relevant problem that better mirrors real-world diagnostic settings, although performance remains modest at this task.
Prior neuroimaging studies have predominantly focused on binary discrimination tasks: distinguishing patients with a specific headache type from healthy controls.

\section{Limitations}

% Explain when your approach may not apply, or things you could not
% check.  \emph{Discussing limitations is essential.  Both ACs and
%   reviewers have been advised to be skeptical of any work that does
%   not consider limitations.}

% Discuss ethical/clinical deployment: require prospective validation, calibration in new populations, and clinician-in-the-loop use.

There are limitations to the present work that should be discussed.
First, our cohort size was relatively small for ML, particularly for subtype analysis, or to perform an comprehensive hyperparameter tuning for the baseline models. 
To partly offset this, we used a lower-dimensional ROI representation for whole-brain analyses rather than a larger atlas with hundreds of regions.
We chose to evaluate model performance using a whole-brain approach rather than seed-based ROI analyses that could further reduce dimensionality.
Second, the headache cohort included participants with different headache diagnoses. This heterogeneity is clinically realistic, but it may also have dampened model performance by reducing label homogeneity.
Third, although the foundation approach achieves strong performance, it offers limited mechanistic interpretability compared to the lower-performing approach based on functional connectivity matrices. 
Lastly, the current work used only rs-fMRI.
This modality is intended to capture intrinsic brain-network organization rather than task-evoked responses; however, uncontrolled variation in participants' internal mental state during scanning may have introduced noise that we could not measure directly.
Promising next steps include leveraging larger cohorts, conducting prospective validation to support clinical translation, and fine-tuning the foundation model while developing more informative attribution analyses.

% ----- end of the main text

\paragraph{Declaration of competing interest}

% All authors must disclose any financial and personal relationships with other people or organizations that could inappropriately influence or bias their work. Examples of potential competing interests include:

The authors declare that they have no known competing financial
interests or personal relationships that could have appeared to influence
the work reported in this paper.

\paragraph{Data availability}

% To foster transparency, you are required to state the availability of any data at submission.

Study data was extracted from the BCH's electronic health record and cannot be shared publicly to protect the privacy of the participants. 
However, it can be shared upon request and subject to relevant approvals.
All our source code is available at \url{https://github.com/cavalab/pediatric-headache-foundation-models}.

\acks{
We would like to thank study physicians at Boston Children’s Hospital Department of Sports Medicine, as well as Dr. Alicia Johnston and Dr. Catherine Lachenauer for their assistance with recruitment.
}

\section*{Funding Support}

% Authors must disclose any funding sources who provided financial support for the conduct of the research and/or preparation of the article. The role of sponsors, if any, should be declared in relation to the study design, collection, analysis and interpretation of data, writing of the report and decision to submit the article for publication. If funding sources had no such involvement this should be stated in your submission.
This work was supported in part by Boston Children's Hospital (BCH) and the National Institutes of Health (NIH) Grant Number R01NS125265.
The neuroimaging research was supported by the Office of the Director, National Institutes of Health under Award Number S10OD025111.
G.S.I.A.~is supported by Coordena\c{c}\~{a}o de Aperfei\c{c}oamento de Pessoal de N\'{i}vel Superior (CAPES) fellowship Finance Code 001.

%Do NOT change font size of references or modify the bibliography style
\bibliography{refs}

\newpage
\appendix
\section{Additional Experimental Details}~\label{apx:methods}

% Some more details about those methods, so we can actually reproduce them. 

This appendix reports the different parameter values tested for each model in preliminary experiments, as well as detailed information about data preparation before generating the different dataset representations.

\subsection{Data preprocessing}

First, the fMRI data were processed using the CONN toolbox and mapped to MNI space. The data obtained after this preprocessing stage were used to generate the functional connectivity matrices.

The NeuroSTORM documentation states that fMRI data should be preprocessed and aligned to MNI152 using a preprocessing tool, which was performed in the previous step. We then adopted the preprocessing pipeline provided by NeuroSTORM by applying spatial interpolation to dimensions smaller than the required size and center-cropping dimensions larger than $96$ voxels.

A z-score normalization was then applied after masking out-of-brain regions. The mask was also used to completely nullify out-of-brain regions, avoiding additional noise from areas irrelevant to the task.

Finally, the temporal dimension was reduced to $20$ non-overlapping frames using average pooling.

\subsection{Model Configurations}

As stated in \S\ref{subsec:ml}, given the low sample size relative to the number of features, as well as the use of bootstrapping and LOO cross-validation, extensive hyperparameter tuning could easily lead to overfitting.
Therefore, we intentionally avoided extensive hyperparameter optimization and restricted tuning primarily to parameters that directly affect model complexity.
Table \ref{tab:ml_hyperparameters} reports the parameters used for the ML models.
Parameters not listed were left at their library defaults using scikit-learn v1.8.0.
During the final experiments, training metrics and SHAP values were recorded for model inspection.

\begin{table}[htb]
    \caption{Parameters of the baseline machine learning models.
    Values in brackets represents different tested configurations.}
    \label{tab:ml_hyperparameters}
    \centering
    \begin{tabular}{lrl}
        \toprule\midrule
        Model & Parameter & Value \\
        \midrule
        Logistic Regression L1 (LR L1) & l1\_ratio & 1.0 (equivalent to penalty='l1')\\
        & solver & saga \\
        & max\_iter & \{100, 1000, 5000\} \\
        \midrule
        Logistic Regression L2 (LR L2) & l1\_ratio & 0.0 (equivalent to penalty='l2')\\
        & solver & liblinear \\
        & max\_iter & \{100, 1000, 5000\} \\ 
        \midrule
        Decision Tree Classifier (DT) & max\_depth & \{None, 2, 3, 5\} \\
        \midrule
        Gradient Boosting Classifier (GB) & n\_estimators & \{5, 100\} \\
        & max\_depth & \{3, default\} \\
        & min\_samples\_leaf & \{10, default\} \\
        \midrule
        Random Forest Classifier (RF) & n\_estimators & \{5, 100\} \\
        & max\_depth & \{None, 3\} \\
        & min\_samples\_leaf & \{10, default\} \\
        \midrule\bottomrule
    \end{tabular}
\end{table}

The ML model using the embedding representation consisted of Logistic Regression with L1 regularization, whose performance was higher in terms of AUROC when trained with data obtained from the functional connectivity representation.

\subsection{Finetuning NeuroSTORM}

\begin{figure}[htb]
    \centering
    \includegraphics[trim={0.71cm 0 0.71cm 0},clip=true,width=\linewidth]{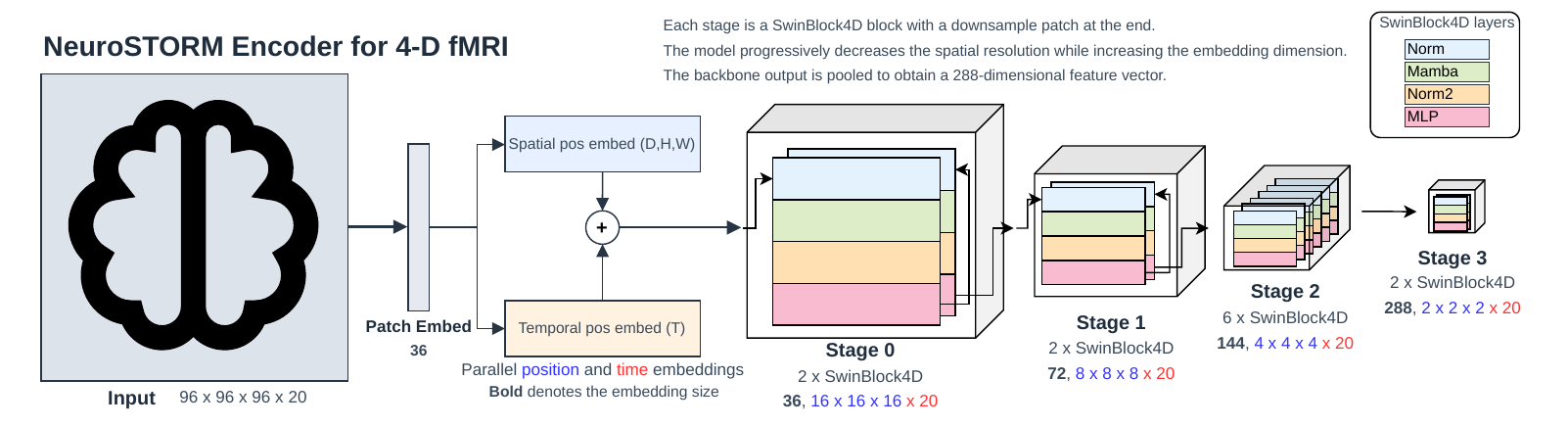}
    \caption{Architecture of the NeuroSTORM encoder backbone.
    A preprocessed 4D fMRI sequence ($96 \times 96 \times 96 \times 20$) is first partitioned into non-overlapping spatial patches and projected into an embedding space. The resulting features are processed through four hierarchical Mamba-based blocks, where the spatial resolution decreases ($16^3 \rightarrow 8^3 \rightarrow 4^3 \rightarrow 2^3$) while the embedding dimension increases ($36 \rightarrow 72 \rightarrow 144 \rightarrow 288$).
    Global pooling produces a 288-dimensional feature representation for downstream prediction.
    }
    \label{fig:neurostorm_arch}
\end{figure}

We performed an exploratory fine-tuning experiment with NeuroSTORM by unfreezing different combinations of the \texttt{norm2} layers in the encoder architecture and combining them with a classification head. We illustrate the NeuroSTORM encoder architecture in Figure \ref{fig:neurostorm_arch}.

First, we performed the training of NeuroSTORM with a classification head using the fixed parameters reported in Table \ref{tab:neurostorm_hyperparameters} and unfreezing every possible combination of the \texttt{norm2} layers at the end of each block (4 blocks in total; 16 possible configurations).

\begin{table}[htb]
    \caption{Parameters of the NeuroSTORM classifier head.}
    \label{tab:neurostorm_hyperparameters}
    \centering
    \begin{tabular}{rl}
        \toprule\midrule
        Parameter & Value \\
        \midrule
        hidden\_dim & 68 \\
        dropout & 0.1 \\
        learning rate & $1 \times 10^{-3}$ \\
        weight decay & $1 \times 10^{-4}$ \\
        batch size & 16 \\
        epochs & 100 \\
        optimizer & AdamW \\
        feature scaling & StandardScaler \\
        loss function & BCEWithLogitsLoss (binary) / \\
        & CrossEntropyLoss (multiclass) \\
        \midrule\bottomrule
    \end{tabular}
\end{table}

The setting that yielded the highest performance in terms of AUROC was unfreezing only the final layer of the initial block, where spatial reduction is minimal.
We then bootstrapped the predictions of the best performing configuration, achieving an AUROC of $0.74\pm0.04$ and an AUPRC of $0.91\pm0.02$, outperforming the demographic model but underperforming a linear classifier trained on NeuroSTORM embeddings.
We found that this fine-tuning approach actually decreased performance, which could be due to the small sample size inherent to this task.
We expect that, given the complexity of pediatric headache, meaningful gains may require fine-tuning on larger pediatric-specific datasets.

\end{document}